\documentclass[sigconf]{acmart}

\usepackage{longfbox}
\usepackage{graphicx}
\usepackage{subfigure}
\usepackage{amsmath}
\usepackage[american]{babel}
\usepackage{microtype}
\usepackage{subcaption}
\usepackage[most]{tcolorbox}
\usepackage{xcolor}
\usepackage{balance}
\usepackage{enumitem} 
\usepackage{color}
\usepackage{url}
\usepackage{arydshln}
\usepackage{bm}
\usepackage{euscript}
\usepackage{mathrsfs}
\usepackage{multirow}
\usepackage{colortbl}
\usepackage{fancyhdr}
\usepackage{graphicx}
\usepackage{caption}
\usepackage{tabularray}
\usepackage{xspace}
\usepackage{makecell}
\usepackage{multirow}
\usepackage{float}
\usepackage{stfloats}

\usepackage{algorithm}
\usepackage{algpseudocode}
\usepackage{wrapfig,booktabs}
\usepackage{amssymb}

\newcommand{\dataset}{WebCode2M\xspace}%
\newcommand{\datasetnospace}{WebCode2M}
\newcommand{\modelname}{WebCoder\xspace}%
\newcommand{\modelnamenospace}{WebCoder}%
\newcommand{\homepage}{\url{https://webcode2m.github.io}}%

\newcommand\mypara[1]{\vspace{1mm}\noindent \textbf{#1} \xspace}

\AtBeginDocument{%
  }

\copyrightyear{2025}
\acmYear{2025}
\setcopyright{acmlicensed}
\acmConference[WWW '25]{Proceedings of the ACM
Web Conference 2025}{April 28-May 2, 2025}{Sydney, NSW, Australia}
\acmBooktitle{Proceedings of the ACM Web Conference 2025 (WWW '25), April
28-May 2, 2025, Sydney, NSW, Australia}
\acmDOI{10.1145/3696410.3714889}
\acmISBN{979-8-4007-1274-6/25/04}

\author{Yi Gui}
\authornote{Both authors contribute equally to this research.}
\orcid{0009-0006-2841-7942}
\affiliation{%
  \institution{Huazhong University of Science and Technology}
  \city{Wuhan}
  \country{China}
}
\additionalaffiliation{%
\department{National Engineering Research Center for Big Data Technology and System, Services Computing Technology and System Lab, Cluster and Grid Computing Lab}
\department{School of Computer Science and Technology}
  \city{Wuhan}
  \state{Hubei}
  \country{China}}

\author{Zhen Li}
\authornotemark[1]
\authornotemark[2]
\orcid{0009-0007-0873-6126}
\affiliation{%
   \institution{Huazhong University of Science and Technology}
  \city{Wuhan}
  \country{China}
}

\author{Yao Wan}
\authornote{Yao Wan is the corresponding author (\texttt{wanyao@hust.edu.cn}).}
\authornotemark[2]
\orcid{0000-0001-6937-4180}
\affiliation{%
   \institution{Huazhong University of Science and Technology}
  \city{Wuhan}
  \country{China}
}

\author{Yemin Shi}
\orcid{0000-0001-9024-7266}
\affiliation{%
  \institution{Peking University}
  \city{Beijing}
  \country{China}
}

\author{Hongyu Zhang}
\orcid{0000-0002-3063-9425}
\affiliation{%
  \institution{Chongqing University}
  \city{Chongqing}
  \country{China}
}

\author{Yi Su}
\orcid{0009-0007-9336-3562}
\affiliation{%
  \institution{Hubei University of Automotive Technology}
  \city{Shiyan}
  \country{China}
}

\author{Bohua Chen}
\orcid{0009-0004-8927-8902}
\affiliation{%
   \institution{Huazhong University of Science and Technology}
  \city{Wuhan}
  \country{China}
}

\author{Dongping Chen}
\orcid{0009-0009-9848-2557}
\affiliation{%
   \institution{Huazhong University of Science and Technology}
  \city{Wuhan}
  \country{China}
}

\author{Siyuan Wu}
\orcid{0009-0006-0960-1699}
\affiliation{%
   \institution{Huazhong University of Science and Technology}
  \city{Wuhan}
  \country{China}
}

\author{Xing Zhou}
\orcid{0009-0006-4165-7756}
\affiliation{%
  \institution{Rabbitpre AI}
  \city{Shenzhen}
  \country{China}
}

\author{Wenbin Jiang}
\authornotemark[2]
\orcid{0000-0001-5628-8806}
\affiliation{%
   \institution{Huazhong University of Science and Technology}
  \city{Wuhan}
  \country{China}
}

\author{Hai Jin}
\authornotemark[2]
\orcid{0000-0002-9254-566X}
\affiliation{%
   \institution{Huazhong University of Science and Technology}
  \city{Wuhan}
  \country{China}
}

\author{Xiangliang Zhang}
\orcid{0000-0002-3574-5665}
\affiliation{%
  \institution{University of Notre Dame}
  \city{Notre Dame}
  \country{United States}
}

\renewcommand{\shortauthors}{Yi Gui et al.}

\begin{document}
\title{\dataset: A Real-World Dataset for Code Generation from Webpage Designs}

\begin{abstract}
Automatically generating webpage code from webpage designs can significantly reduce the workload of front-end developers, and recent \textit{Multimodal Large Language Models} (MLLMs) have shown promising potential in this area. 
However, our investigation reveals that most existing MLLMs are constrained by the absence of high-quality, large-scale, real-world datasets, resulting in inadequate performance in automated webpage code generation.
To fill this gap, this paper introduces \datasetnospace, a new dataset comprising 2.56 million instances, each containing a design image along with the corresponding webpage code and layout details. Sourced from real-world web resources, \dataset offers a rich and valuable dataset for webpage code generation across a variety of applications.
The dataset quality is ensured by a scoring model that filters out instances with aesthetic deficiencies or other incomplete elements. 
To validate the effectiveness of \datasetnospace, we introduce a baseline model based on the \textit{Vision Transformer} (ViT), named \modelname, and establish a benchmark for fair comparison.
Additionally, we introduce a new metric, TreeBLEU, to measure the structural hierarchy recall.
The benchmarking results demonstrate that our dataset significantly improves the ability of MLLMs to generate code from webpage designs, confirming its effectiveness and usability for future applications in front-end design tools.
Finally, we highlight several practical challenges introduced by our dataset, calling for further research.
The code and dataset are publicly available at our project homepage: \homepage.
\end{abstract}



\maketitle

\section{Introduction}
Front-end software developers typically create webpages based on \textit{Graphical User Interface} (GUI) mockups designed by UI designers. However, this process is often time-consuming and costly. 
To this end, several neural models have been proposed to automate the process of generating code from GUI design images, thereby alleviating the burden on front-end developers.
Among these, pix2code~\cite{Tony2018_pix2code} and sketch2code~\cite{Alex2019_Sketch2code} are two exemplary works that translate images, whether simple-styled UI designs or hand-drawn sketches, into front-end code.
Recently, \textit{Multimodal Large Language Models} (MLLMs), such as GPT-4V~\cite{Ouyang2022TrainingLM}, have also demonstrated impressive potential in this area.

Despite its potential, we are still far from fully automating front-end engineering to achieve true ``\textit{screenshot in, code out}'' functionality. 
In particular, as highlighted in a recent study~\cite{Si2024Design2CodeHF}, the complexity of code generation increases with the growth in the total number of \textit{HyperText Markup Language} (HTML) tags, the diversity of unique tags, and the depth of the \textit{Document Object Model} (DOM) tree. Modern MLLMs, such as GPT-4V, also exhibit a notable decline in performance when confronted with real-world webpage designs that feature complex structures and a larger variety of unique HTML tags~\cite{Si2024Design2CodeHF}.

One possible solution lies in fine-tuning pre-trained LLMs, with the potential for improved performance as the amount of data increases. However, this approach faces a significant limitation since existing datasets are either too small to provide meaningful generalization~\cite{Si2024Design2CodeHF, iwbench_guo_2024} or consist of synthetic data that does not fully capture the complexity and variability of real-world webpage designs~\cite{Laurenccon2024UnlockingTC, DBLP:journals/corr/abs-2406-20098}. For instance, Design2Code~\cite{Si2024Design2CodeHF} contains only 484 real-world samples, intended solely for testing and insufficient for effective fine-tuning.
WebSight~\cite{Laurenccon2024UnlockingTC} is another dataset comprising approximately 0.8 million synthesized samples generated by LLMs. However, a significant disparity exists between these samples and real-world data~\cite{Si2024Design2CodeHF}.
Specifically, WebSight samples average 647 tokens, 19 tags, and a DOM depth of 5, whereas our study reveals that real-world samples can involve up to 50 times more tokens, six times as many tags, and double the DOM depth (see Figure~\ref{fig_samples}). 
This substantial gap between synthetic and real-world data can limit the practical effectiveness of fine-tuned MLLMs when applied to more complex, real-world scenarios.

\mypara{Our Work.}
To fill this gap, this paper introduces a large-scale real-world dataset for webpage generation, named \datasetnospace, which includes 2.56 million instances. 
Each instance features a high-quality webpage design image paired with its corresponding HTML and \textit{Cascading Style Sheets} (CSS) code. This dataset overcomes the limitations of existing datasets by offering a diverse and comprehensive collection of real-world webpage designs and their associated code. On average, the samples contain 31,216 tokens, 158 tags, and a DOM depth of 13. \dataset is poised to be an invaluable resource for advancing the development of webpage code generation models.

To construct our dataset, we first collect approximately 0.5 billion samples from the Common Crawl dataset~\cite{ccdataset}, which includes a diverse array of web domains and styles. For each webpage, we extract the associated CSS code and image elements, remove noise and irrelevant code, and generate screenshots. To ensure data quality, we develop a scoring model to filter out instances with incomplete elements or suboptimal aesthetic quality, such as disorganized layouts or excessive blank spaces. 
This scoring model is trained on a manually annotated subset of 10,000 entries, curated by six annotators using consensus-based annotation, achieving a validation accuracy of 90\% in distinguishing high-quality instances from low-quality ones.
\begin{figure*}[t]
    \centering
    \includegraphics[width=0.98\linewidth]{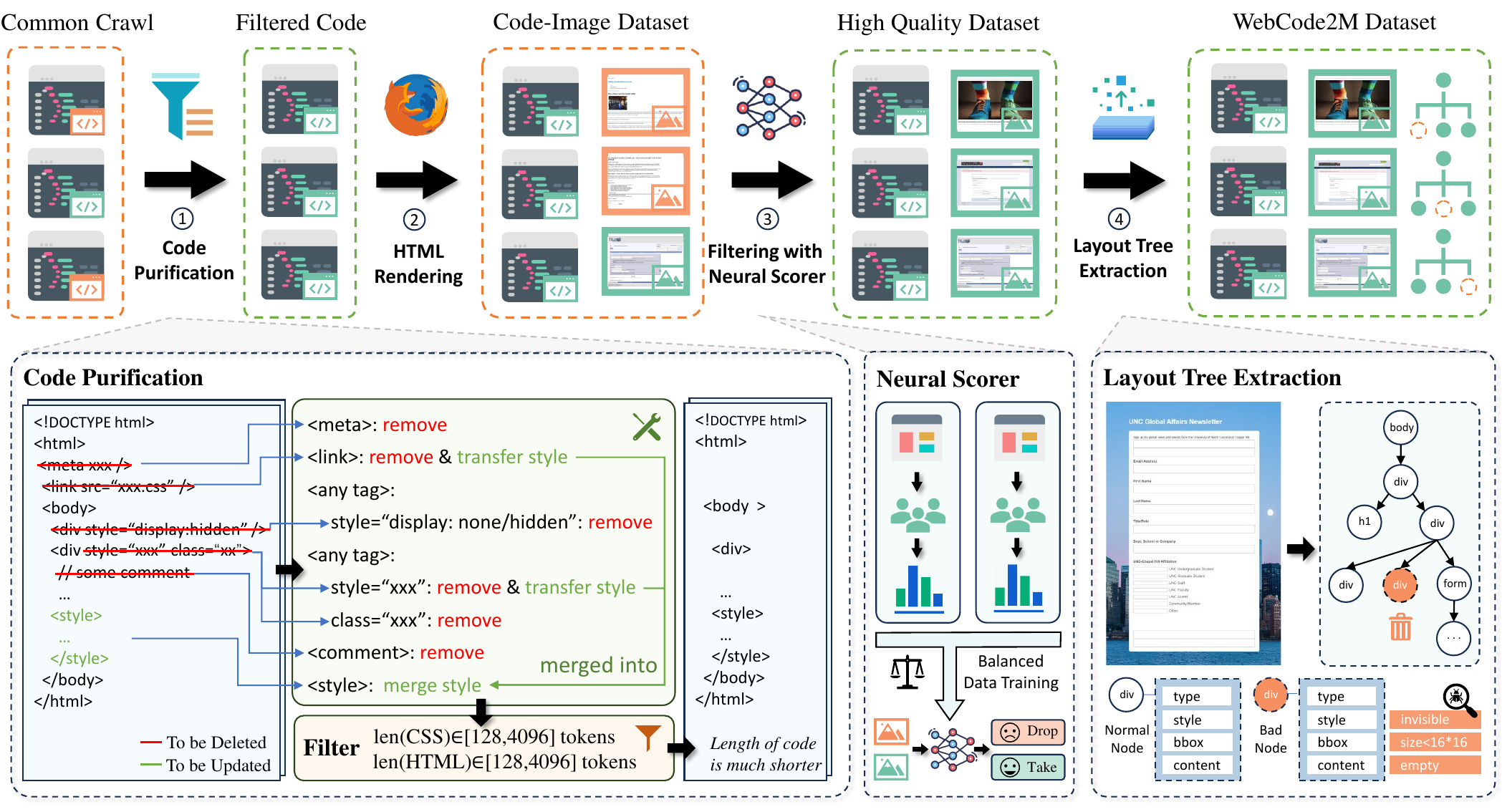}    
    \caption{The pipeline of constructing the \dataset dataset.}
    \label{fig_pipeline} 
\end{figure*}

To demonstrate the potential of our dataset for improving automatic webpage generation, we fine-tune a ViT model~\cite{alexey2020image} as a new baseline, termed \modelname, for translating webpage design images into HTML/CSS code and establish a benchmark for fair comparison.
Compared to the two fine-tuned baselines, Design2Code-18B~\cite{Si2024Design2CodeHF} and WebSight VLM-8B~\cite{Laurenccon2024UnlockingTC}, our model, fine-tuned from the smaller Pix2Struct-1.3B, outperforms both across all evaluation metrics, including CLIP-based visual similarity~\cite{DBLP:conf/icml/RadfordKHRGASAM21}, low-level appearance accuracy~\cite{Si2024Design2CodeHF}, and our proposed TreeBLEU to measure the structural hierarchy recall.
We also benchmark a broad array of general-purpose MLLMs, including the LLaVA family~\cite{liu2023llava}, CogAgent-Chat-18B~\cite{DBLP:journals/corr/abs-2312-08914}, GPT-4V, GPT-4o~\cite{DBLP:journals/corr/abs-2303-08774}, Gemini~\cite{DBLP:journals/corr/abs-2312-11805}, and Claude~\cite{TheC3}. Experimental results show that \modelname outperforms these models across most evaluation metrics. The only exception is GPT-4o, which achieves higher similarity in CLIP and visual appearance but has a lower substructure recall rate.

\mypara{Contributions.}
The contributions of this paper are as follows.
\begin{itemize}[leftmargin=*]
    \item \textbf{New Dataset.} To the best of our knowledge, \dataset is the first real-world and large-scale dataset tailored to empower MLLMs in the domain of generating webpage code from high-fidelity images.
    \item \textbf{Comprehensive Benchmark.} We fine-tune a ViT model, named \modelname, on our \dataset dataset and evaluate it through a comprehensive set of experiments alongside other fine-tuned baselines. Experimental results demonstrate the effectiveness of the dataset in enabling MLLMs to automatically generate code from webpage designs.
    Additionally, we introduce a novel metric, TreeBLEU, to measure the structural hierarchy recall.
    \item \textbf{Open-Source Resources.} We open-source the code base, the dataset, and the new benchmark model, making them freely available to the research and developer communities,  for further innovation in automating front-end engineering. The resources are available at \homepage.    
\end{itemize}

\section{\dataset: The Dataset}

\subsection{Dataset Construction}
The aim of this study is to curate a dataset that facilitates training neural models to generate code from webpage designs.
As large-scale human-designed screenshots are hard to collect manually, we opt to reversely generate screenshot image from a curated open-source web dataset via rendering the webpage code.
Figure~\ref{fig_pipeline} illustrates the pipeline for constructing \dataset, encompassing steps such as code purification, HTML rendering, filtering with a neural scorer, and layout tree extraction.

\mypara{Raw Data Collection.}
We build our dataset on top of the Common Crawl dataset~\cite{ccdataset}, a comprehensive collection of global webpage data spanning from 2013 to the present, updated monthly through web crawling. Previously, the Common Crawl dataset is primarily used for pre-training models on text-based tasks. Due to our computational resource and download speed constraints, we randomly sample approximately 0.5 billion webpages from the first segment of the \texttt{CC-MAIN-2023-50} version, which contains about 3.35 billion webpages, as our initial data. 

\mypara{Code Purification.}
Our investigation reveals that webpage code from online websites is often lengthy and includes redundant elements, which significantly impairs the model’s ability to accurately learn the correlation between webpage code and screenshots~\cite{Laurenccon2024UnlockingTC}.
To ensure data quality, we meticulously clean the combined HTML and CSS text adhering to the following steps.
\begin{itemize}[leftmargin=*]
    \item 
    \underline{Quick length filtering.}
    We filter out samples that are either excessively long or too short. 
    This is because parsing errors from Python tools often lead to excessively short HTML or CSS code, while excessively long input significantly slows down our training and inference procedures.
    Specifically, we employ a rapid filtering method based on code length, measured by the number of characters. Assuming that one word is approximately five characters long, we establish length ranges for HTML and CSS code between $[128\times5$, $2048\times5]$ characters and $[128\times5, 4096\times5]$ characters, respectively. Webpages that fall outside these ranges are filtered out.
    \item 
    \underline{Redundant code elements cleansing.}
    The code samples in the raw dataset may include redundant elements, such as comments and hidden elements, as well as components that do not directly affect the rendering of static HTML pages. To address this, we propose removing the following contents from both HTML and CSS code: comments, \texttt{<meta>} and \texttt{<script>} tags, hidden elements (hidden, zero-sized, or outside the display range), attributes not in (\texttt{class}, \texttt{id}, \texttt{width}, \texttt{height}, \texttt{style}, \texttt{src}) of all HTML elements, and CSS styles that are not effective in the HTML code.
    In addition, we transfer and merge all the external and inline CSS code into a single \texttt{<style>} tag.
\end{itemize}

\begin{figure*}[t]
    \centering
    \includegraphics[width=\linewidth]{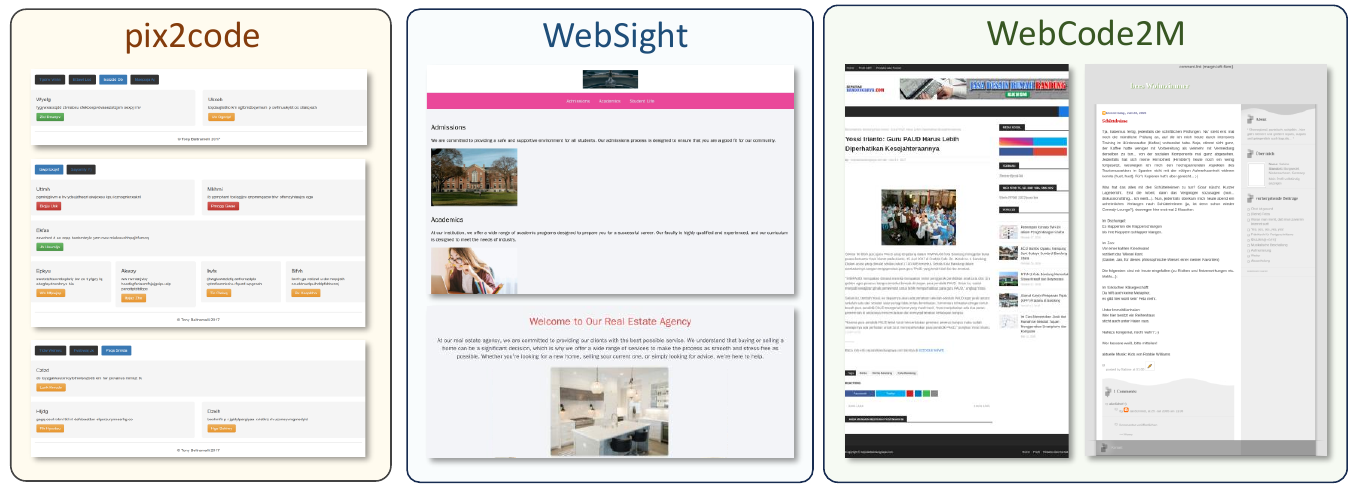}
    \caption{Representative screenshots of webpages in \dataset and other datasets.}
    \label{fig_samples}
\end{figure*}

\mypara{HTML Rendering for Screenshot Generation.} 
After purifying the data, we generate webpage screenshots from the combined HTML and CSS code. This process is implemented using Playwright~\cite{homepage_Playwright}, a headless browser automation tool that allows us to render webpages and capture high-fidelity screenshots. By simulating a real browser environment, Playwright ensures that the rendered webpage accurately reflects the appearance of the HTML and CSS code. 
This process is highly time-consuming, accounting for roughly 80\% of the total processing time, which spans approximately one month.

\mypara{Filtering with a Neural Scorer.}\label{sec_scorer}
In our empirical data analysis, we observe that a considerable proportion of the generated screenshots exhibit deficiencies in aesthetics. These low-quality screenshots are generally attributed to incompletely loaded pages resulted from various factors, for instance, invalid image links, and cases where the content is mainly composed of textual content.

The presence of flawed screenshots can compromise the overall quality of the dataset, necessitating a rigorous filtering of the acquired data. Given the large volume of our dataset, manually screening all the data is impractical. Therefore, we train a classification model to serve as a neural scorer, assessing the screenshots and subsequently eliminating samples that fall below a specified score threshold.

In practice, we devise an annotating tool and manually annotate a subset of the generated screenshots. The scoring criteria are thoughtfully crafted, and each criterion satisfied will be awarded one point:
(1) Normal webpage layout (human-designed layout, not simple auto single-column arrangement);
(2) Normal webpage styling (elements like lists and blocks are styled, not using default styles);
(3) No excessive blank areas;
(4) Rich color combinations; and
(5) Good aesthetic appearance.
During the manual annotation process, we invite six annotators who hold a Bachelor's degree in Computer Science and have at least three years of web development experience. We then divide them into two groups to perform consensus annotation, where annotators within each group evaluate the same data.
This annotation strategy minimizes the influence of subjective factors on the scoring results.
The annotation process takes approximately two weeks for all the participants, ultimately yielding 10,000 manually scored data entries.

\begin{figure}[t]
    \centering
    \includegraphics[width=\linewidth]{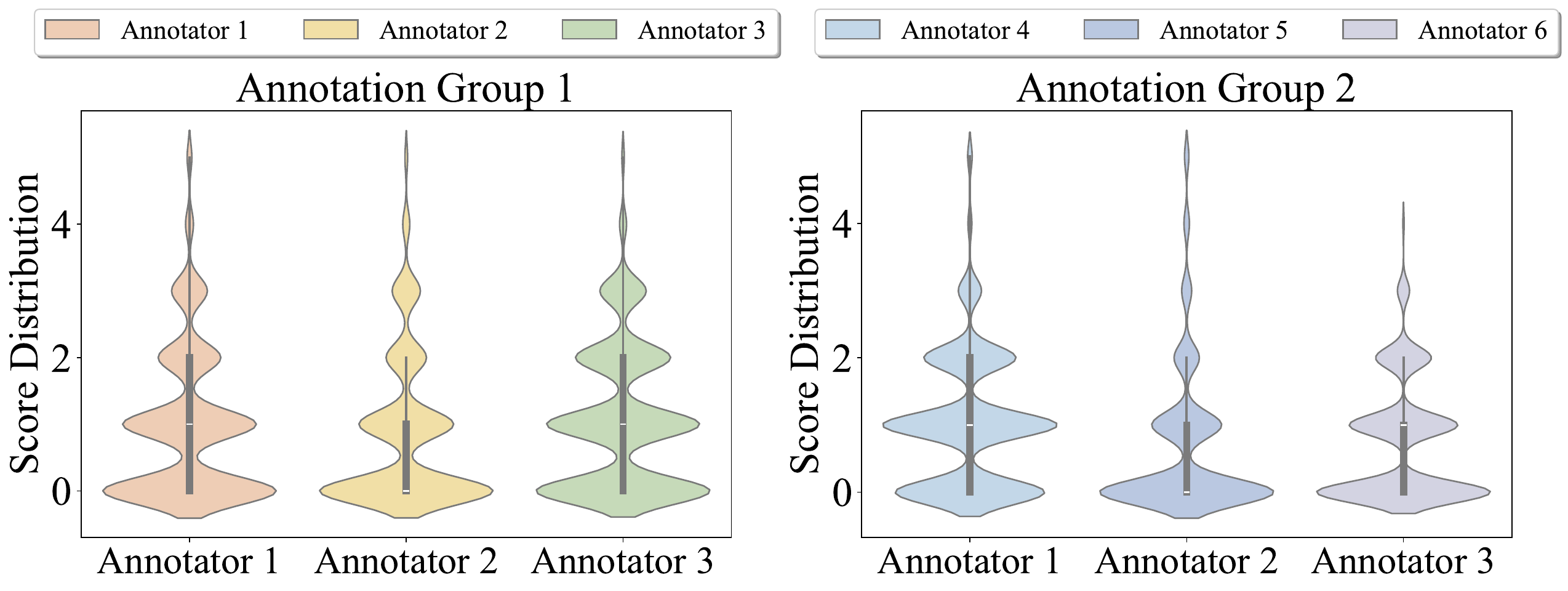} 
    \caption{Score distributions of annotators in two groups.}
    \label{fig_annotation_consistency}
\end{figure}

\begin{figure}[t]
\includegraphics[width=0.7\linewidth]{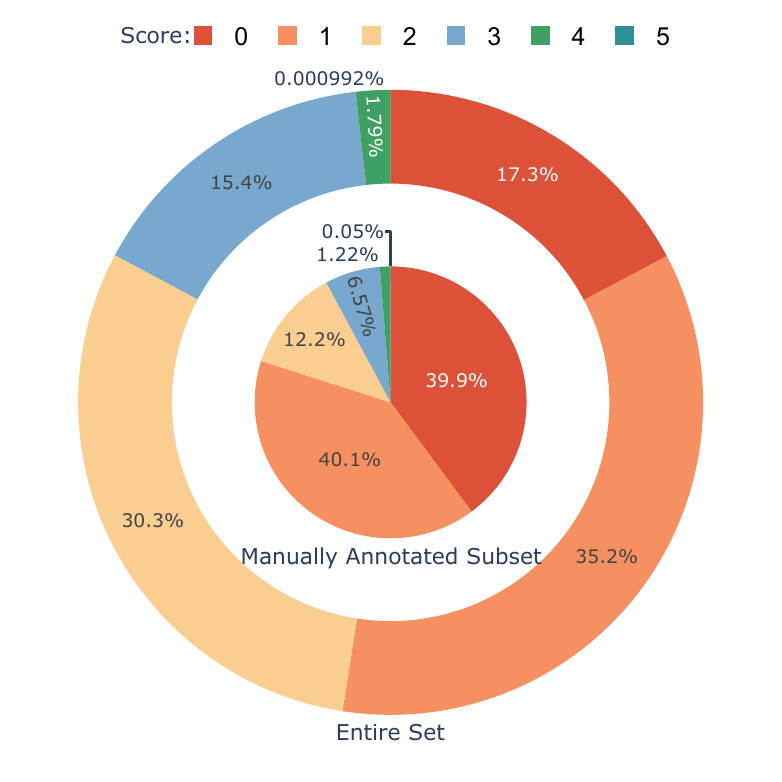} 
        \captionof{figure}{Score distribution of the manually annotated subset (inner ring) and the entire dataset (outer ring) before score-based filtering.}
        \label{fig_score}
\end{figure}

The score distribution of the manually labeled subset is depicted in Figure~\ref{fig_score} (inner ring). The statistics reveal that 80\% of the data fell within the low-quality category, scoring between 0 and 1. Conversely, only 20\% of the entries, exhibiting scores of 2 or higher, demonstrate a commendable level of structural integrity and aesthetic appeal.
We also conduct a consistency analysis of all the data from the annotators (see Figure~\ref{fig_annotation_consistency}). Although certain differences exist among annotators within the same group, the overall trend remains similar. By averaging scores within the same group, the impact of subjectivity is significantly reduced.
Utilizing the rated data, we train a \textit{ResNet-50}~\cite{DBLP:conf/cvpr/HeZRS16} model to serve as a scorer, predicting the score of input screenshots. 
This scorer achieves 75\% accuracy on the test subset of the manually scored dataset and nearly 90\% accuracy in binary classification, determining whether the score is greater than or equal to two.
Using this scorer, we remove samples with scores less than two, which accounted for 52.5\% of the entire raw dataset (as shown in the outer circle of Figure~\ref{fig_score}).

\mypara{Layout Tree Extraction.}
Considering that the webpage's layout defines the spatial arrangement and relationships between UI components, it can serve as a critical source of information. If available, the layout can act as a training target for the model, facilitating code generation by guiding the model to understand not only the structure of the webpage but also the precise positioning of elements. Thus, each data instance in our dataset is upgraded to a triplet: (webpage code, design image, layout). The layout, represented by the bounding boxes (BBox) of HTML elements, includes key information such as the size, location, and hierarchy of page components. This additional layout data will aid the model in learning to generate the webpage DOM tree structure more accurately. 

\subsection{Ethical Compliance} 
As our dataset is sourced from online webpages, it may contain content that is inappropriate for public release, such as explicit material or violent content.
To mitigate ethical concerns regarding potential negative impacts, such as the misuse of models trained on this dataset, we perform additional filtering steps. Specifically, we apply an image filter to the screenshots and a profanity filter to the web text. Only samples that passed both filters are retained. Detailed filtering procedures are provided in Appendix~\ref{sec_harmfull_filtering}.
    
\begin{figure}[t]
    \centering
        \centering
        \includegraphics[width=0.72\linewidth]{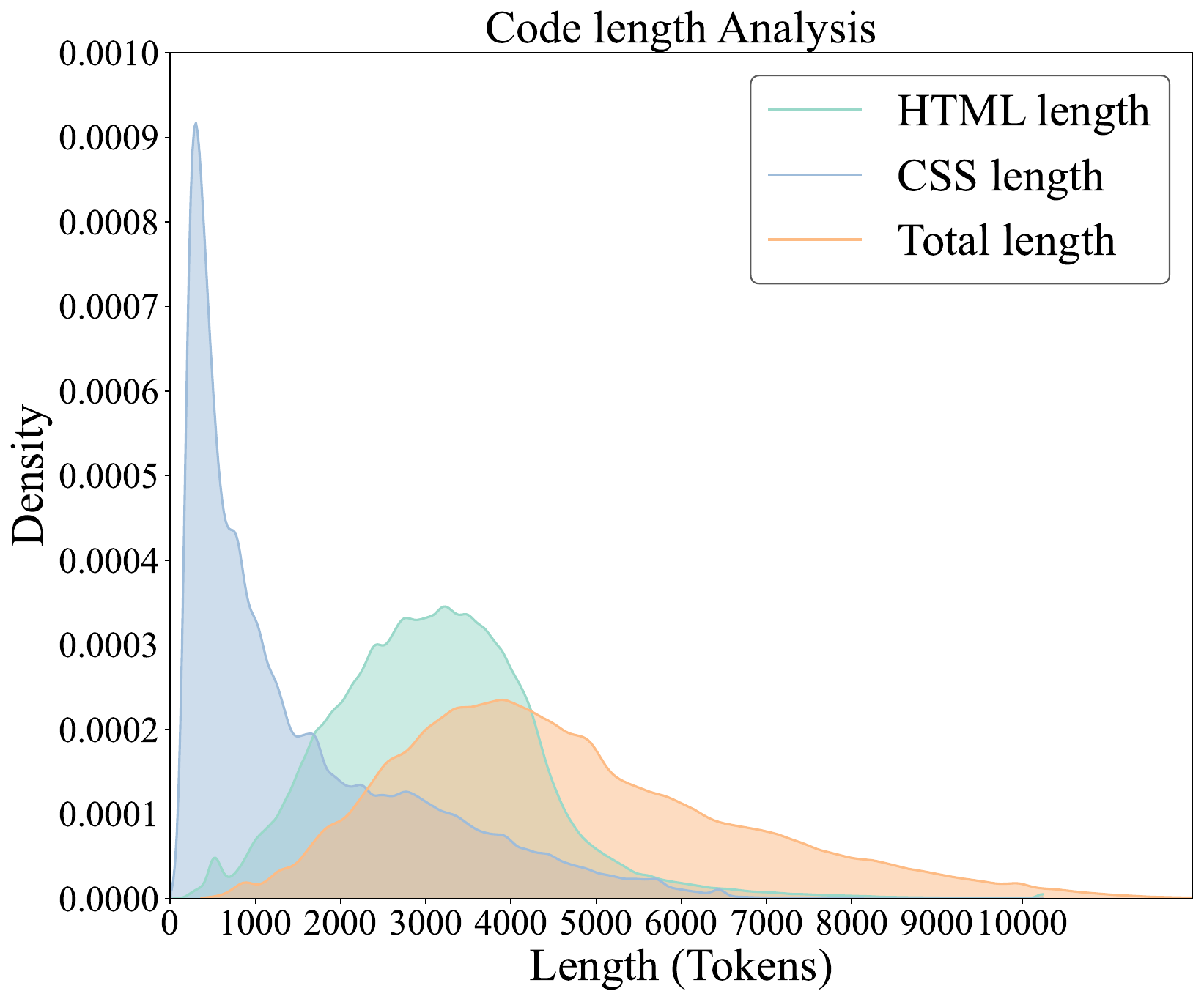} 
        \captionof{figure}{Length density of the \dataset dataset.}
        \label{fig_len}
\end{figure}

\subsection{Dataset De-Duplication and Partition}
To support the use of our dataset in both fine-tuning models for webpage code generation and evaluating their performance, we organize the data into well-structured partitions. After using the hash codes of the screenshots to quickly de-duplicate the refined dataset, which comprises millions of entries, we sample approximately two thousand entries with a score above 4 as our candidate test dataset. The remaining 2.56 million entries serve as the training dataset. For the candidate test dataset, we further remove duplicates using CLIP~\cite{DBLP:conf/icml/RadfordKHRGASAM21} similarity and conduct a manual inspection on each entry.

Furthermore, we partition the test subsets based on code length to assess the model's code generation capability across varying levels of difficulty. As illustrated in Figure~\ref{fig_len}, the dataset shows a wide range of data length variations. Specifically, we use two length thresholds (i.e., 2048 and 4096), to select 256 entries from within three length ranges, thus creating three test subsets. We refer to them as \dataset-Short, \dataset-Mid, and \dataset-Long. 
Table~\ref{tb_subsets_len} summarizes the overall statistics and provides detailed length statistics of \dataset for both training and testing.

\subsection{Dataset Characteristics}
Upon acquiring the final dataset \dataset, we conduct an analysis to identify several key characteristics. To quantitatively assess the diversity and quality of our dataset, we employ the same statistical metrics used in Design2Code, facilitating a comparison with other datasets. The results are presented in Table~\ref{tb_stat}. Specifically, Avg. Len represents the token length as determined by the GPT-2 tokenizer~\cite{hfgpt2}; Avg. Tags indicates the total number of tags in the HTML code; Avg. Unique Tags denotes the count of distinct tags in the HTML code; and Avg. DOM Depth signifies the maximum depth of the HTML's DOM tree.

\begin{table}[!t]
    \centering
    \captionof{table}{Dataset partition.} 
    \begin{tabular}{lccc} 
    \Xhline{1px}
    Subset            &Purpose    & Size      &   Length (Tokens)  \\ 
    \Xhline{0.7px}
    \dataset          &Training    & 2,563,905 & {[}368,16668]         \\
    \dataset-Short &Testing      & 256       & {[}551, 2045]         \\
    \dataset-Mid   &Testing  & 256       & {[}2052, 4085]        \\
    \dataset-Long  &Testing & 256       & {[}4098, 10990]       \\
    \Xhline{1px}
    \end{tabular}    
    \label{tb_subsets_len}
\end{table}

\begin{table*}[t!]
\setlength{\tabcolsep}{2.2pt}
\centering
\caption{A statistical comparison between \dataset and all the publicly available datasets. The statistical data of WebSight and Design2Code is referred to~\cite{Si2024Design2CodeHF}.}
\label{tb_stat}
\begin{tabular}{lccccccc} 
\Xhline{1px}
Dataset             & pix2code      & WebSight  & Design2Code     & \datasetnospace                 & \datasetnospace-Short          & \datasetnospace-Mid            & \datasetnospace-Long            \\ 
\Xhline{0.7px}
Purpose             & Training\&Testing & Training  & Testing         & Training                  & Testing                   & Testing                   & Testing                    \\
Source              & Synthetic     & Synthetic & Real-World & Real-World & Real-World & Real-World & Real-World  \\
Size (\#samples)    & 1742          & 0.8M      & 484             & 2.5M                      & 256                       & 256                       & 256                        \\
Avg. Len (\#tokens) & 1316±177      & 647±216   & 31216±23902     & 5366±2393                 & 2025±514                  & 3750±765                  & 7940±1853                  \\
Avg. Tags           & 52±8          & 19±8      & 158±100         & 184±77                    & 81±34                     & 144±61                    & 222±81                     \\
Avg. DOM Depth      & 8±0           & 5±1       & 13±5            & 15±5                      & 10±4                      & 13±7                      & 16±4                       \\
Avg. Unique Tags    & 17±0          & 10±3      & 22±6            & 24±6                      & 18±4                      & 21±5                      & 26±5                       \\
\Xhline{1px}
\end{tabular}
\end{table*}
\mypara{Superior Diversity.}
From Table~\ref{tb_stat}, it is apparent to see that
our dataset contains a significantly greater number and diversity of HTML tags and exhibits a more intricate DOM tree structure compared to the pix2code and WebSight datasets. This suggests that our dataset, sourced from real-world webpages, offers a remarkable diversity advantage over synthetic datasets generated by LLMs like WebSight. Design2Code, which also utilizes real-world data through the C4 dataset~\cite{DBLP:journals/jmlr/RaffelSRLNMZLL20} from the Common Crawl corpus, exhibits a comparable distribution across these metrics, underscoring the benefits of real-world data in capturing the complexity of actual webpages. Moreover, this comparison highlights significant deviations in data attribute distributions between LLM-generated datasets and real webpages.

Figure~\ref{fig_samples} presents several representative screenshots from the datasets (excluding Design2Code). The pix2code dataset comprises basic block elements and text-based UI elements, suitable for both Android and iOS UIs. In contrast, WebSight consists of structurally simple webpages. Our dataset, on the other hand, closely mirrors typical real webpages, featuring a variety of layout structures and rich elements such as images. Additionally, our dataset captures webpages in a diverse range of languages (see Figure~\ref{fig_lang} in Appendix).

\mypara{Large Scale and High Quality.}
Compared to pix2code and Design2Code, which contain only a few thousand or fewer data samples, and WebSight, which includes 0.8 million samples, our dataset is significantly larger, comprising 2.56 million samples. This dataset includes both a comprehensive training dataset and a high-quality test dataset, making it much larger in scale.
Notably, compared to Design2Code, our \dataset significantly reduces the average code length to about one-tenth of its original size, while maintaining the diversity and quantity of HTML tags, thereby preserving the high quality of the dataset.

\section{Benchmarking}
We introduce a baseline model based on the ViT, named \modelname, and establish a benchmark for fair comparison. 

\subsection{\modelname: A Reference Baseline}
To demonstrate the potential of our dataset in enhancing automatic webpage generation, we fine-tune a ViT model to establish a new baseline for translating design images into HTML/CSS code. Specifically, we select Google's Pix2Struct-1.3B~\cite{Kenton2023_Pix2Struct} as our base model. This model, based on the ViT architecture, has been pre-trained on webpage code derived from URLs in the C4 dataset~\cite{DBLP:journals/jmlr/RaffelSRLNMZLL20}. Pix2Struct-1.3B is notable for its robustness to extreme aspect ratios and ability to adapt dynamically to changes in sequence length and resolution. We conduct a full fine-tuning of the pre-trained Pix2Struct on our training dataset, resulting in our model, \modelname.

\subsection{Setup and Baselines}
Our experiments focus on two primary \textit{Research Questions} (RQs).

\mypara{RQ1: The effectiveness of the training dataset.}  To investigate the ability of our training dataset to empower MLLMs in webpage generation, we compare \modelname\ with several state-of-art models which are also fine-tuned specifically for this task.
\begin{itemize}[leftmargin=*]
\item  \textbf{WebSight VLM-8B~\cite{Laurenccon2024UnlockingTC}.}
Hugging Face utilizes its training dataset (WebSight) and the DoRA~\cite{DBLP:journals/corr/abs-2402-09353} mechanism to fine-tune a base VLM, which has been pre-trained on image/text pairs.
\item  \textbf{Design2Code-18B~\cite{Si2024Design2CodeHF}.}
Stanford's Design2Code is also fine-tuned on the WebSight dataset. It adopts CogAgent~\cite{DBLP:journals/corr/abs-2312-08914} as its base model and utilizes LoRA~\cite{DBLP:conf/iclr/HuSWALWWC22} as the fine-tuning method to accelerate the training process.
\item \textbf{\modelnamenospace*.}
Another Pix2Struct model in the same setting but trained on the WebSight dataset for comparative experiments.
\end{itemize}

\mypara{RQ2: Benchmarking on the test datasets.} We also introduce a broad array of the latest and most powerful general-purpose pre-trained MLLMs for benchmarking.

\begin{itemize}[leftmargin=*]
\item  \textbf{LLaVA Family~\cite{liu2023llava}.}
The LLaVA family consists of various MLLMs that connect a vision encoder and an LLM for general-purpose visual and language understanding.
In our work, we introduce \textbf{LLaVA-v1.5-7B}, \textbf{LLaVA-onevision-0.5B}, and \textbf{LLaVA-onevision-7B} as the baselines. The prompt used for these models follows~\cite{homepage_screen2shot} and is detailed in Figure~\ref{fig_example_for_pormpt}.

\item  \textbf{CogAgent-Chat-18B~\cite{DBLP:journals/corr/abs-2312-08914}.}
CogAgent-Chat-18B is a general MLLM that supports both low- and high-resolution images and performs quite well on webpage navigation.
We also input the screenshot and a simple prompt ``\textit{write an HTML code}'' to generate the webpage as Design2Code does.

\item \textbf{Commercial Models.}
Some general commercial models have demonstrated impressive performance across various fields, proficient in both code generation and web understanding. Therefore, we introduce OpenAI's \textbf{GPT-4V} and \textbf{GPT-4o}~\cite{DBLP:journals/corr/abs-2303-08774}, Google DeepMind's \textbf{Gemini}\footnote{Around one-third of the test cases fail due to the ``recitation'' error of Gemini. }
~\cite{DBLP:journals/corr/abs-2312-11805}, and Anthropic's \textbf{Claude}~\cite{TheC3} as baselines. The prompt for these models also follows~\cite{homepage_screen2shot}, detailed in Figure~\ref{fig_example_for_pormpt}.
Although previous work~\cite{Si2024Design2CodeHF} suggests that multi-round generation methods (\textit{e.g.}, self-correction) may outperform one-pass generation, baseline models such as Design2Code-18B and WebSight VLM-8B are fine-tuned with one-pass dataset and only support one-pass generation. Therefore, to ensure a fair comparison, all baselines will employ the one-pass generation strategy.
\end{itemize}

\subsection{Evaluation Metrics}\label{sec_metrics}

\mypara{Visual Similarity Measurement.}
We adopt CLIP~\cite{DBLP:conf/icml/RadfordKHRGASAM21} similarity and \textit{ Visual Score}~\cite{Si2024Design2CodeHF}\footnote{
We use the Visual Score from the first version of their paper, which may differ from the latest version, particularly with respect to the sub-indicator \textit{text color}.
} as two metrics to assess visual similarities.
CLIP similarity is derived from calculating the cosine value of two images' latent vectors encoded by CLIP, which measures the overall visual similarity.
\textit{Visual Score} is utilized to measure the matching degree of low-level elements in terms of appearance, calculating the average scores of the matching ratio between the reference and candidate blocks, as well as the similarity at four block levels in terms of color, text, CLIP, and position.

\mypara{Structure Similarity Measurement.}
Skeletons of webpage code that determines the layout and appearance of the page, also known as the HTML DOM tree, can also serve as a metric to evaluate structural similarity that compares ground truth (for instance, during the training phase or when the target code is provided in the inference stage) and the DOM tree of the generated code.
Inspired by~\cite{Ren2020CodeBLEUAM}, we propose a new metric \textbf{TreeBLEU} to evaluate the matching degree of the generated HTMLs' DOM tree (without terminal nodes that contain tags' attributes, \textit{e.g.}, content and style) compared to the ground truth.

TreeBLEU is defined as the proportion of all 1-height subtrees~(see Algorithm~\ref{alg_subtrees}) in a given tree that can be matched with that of a reference tree.
Let $S(.)$ be the set of 1-height subtrees, then it can be formulated as:
\begin{equation}
    \operatorname{TreeBLEU}=\frac{{\left| S(t) \cap S(\hat{t}) \right|}}{{\left| S(\hat{t}) \right|}}\,,
    \label{eq_treebleu}
\end{equation}   
where $t$ and $\hat{t}$ denote the given and reference trees, respectively. 

\subsection{Implementation Details}
We configure the model to handle up to 1,024 image patches and set the maximum sequence length to 2,048 tokens. Due to GPU memory constraints, we use a batch size of 1 during training.
Initially, we fine-tune the model on a subset of the \dataset dataset with a sequence length of 2,048 tokens for three epochs (90,000 iterations), using a maximum learning rate of 5$e$-5 and a cosine scheduler. This phase aims to enable code generation from visual inputs.
Next, we refine the model with a lower learning rate (1$e$-5) over three more epochs on another dataset subset, reducing the sequence length to 1,024 tokens across 10,000 iterations.
All the experiments are run on a Linux server equipped with 4 NVIDIA A800 80GB GPUs.

\begin{table*} 
\caption{The performance comparison among the specialized models (the best is marked in bold).} 
\centering
\setlength{\tabcolsep}{4.2pt} 
\begin{tabular}{lcccccccccc} 
\Xhline{1px}
\multirow{2}{*}{\textbf{Model}}   & \multirow{2}{*}{\textbf{Training Dataset}}   &  \multicolumn{3}{c}{\makebox[0pt][c]{\hspace{1.6em}\textbf{\datasetnospace-Short}\hspace{-10.1em}\rule[-0.7ex]{11.8em}{0.5px}}}   & \multicolumn{3}{c}{\textbf{\makebox[0pt][c]{\hspace{1.6em}\textbf{\datasetnospace-Mid}\hspace{-9.5em}\rule[-0.7ex]{11.8em}{0.5px}}}}  & \multicolumn{3}{c}{\makebox[0pt][c]{\hspace{1.6em}\textbf{\datasetnospace-Long}\hspace{-9.8em}\rule[-0.7ex]{11.8em}{0.5px}}} \\
                                     
&                &  Visual         & CLIP          & TreeBLEU        &Visual           & CLIP         &  TreeBLEU       &  Visual     & CLIP         &   TreeBLEU       \\ 
\Xhline{0.7px}
WebSight VLM-7B  & WebSight      & $.57_{\pm.24}$  & $.69_{\pm.12}$ & $.03_{\pm.04}$  & $.52_{\pm.23}$  & $.67_{\pm.11}$ & $.03_{\pm.04}$  & $.48_{\pm.27}$ & $.64_{\pm.11}$ & $.03_{\pm.03}$  \\
Design2Code-18B  & WebSight      & $.75_{\pm.14}$  & $.68_{\pm.10}$ & $.04_{\pm.05}$  & $.69_{\pm.23}$  &$.70_{\pm.10}$  & $.05_{\pm.05}$   & $.61_{\pm.28}$ & $.68_{\pm.10}$  & $.06_{\pm.03}$  \\
\modelname*-1.3B & WebSight      & $.42_{\pm.32}$  & $.68_{\pm.11}$ & $.06_{\pm.06}$  & $.36_{\pm.30}$   & $.67_{\pm.11}$ & $.04_{\pm.04}$   & $.38_{\pm.29}$ & $.65_{\pm.11}$  & $.04_{\pm.04}$  \\
\rowcolor{gray!10!white} \modelname-1.3B & \datasetnospace     & $\bm{.78_{\pm.25}}$  & $\bm{.73_{\pm.13}}$ & $\bm{.35_{\pm.17}}$ & $\bm{.69_{\pm.19}}$  & $\bm{.71_{\pm.10}}$ & $\bm{.22_{\pm.11}}$  & $\bm{.65_{\pm.21}}$ & $\bm{.69_{\pm.12}}$  & $\bm{.15_{\pm.07}}$ \\
\Xhline{1px}
\end{tabular}
\label{tb_results_1}
\end{table*}

\begin{table*} 
\caption{Benchmarking performance of several general-purpose
MLLMs using the \dataset (the best is marked in bold).} 
\centering
\setlength{\tabcolsep}{7.5pt} 
\begin{tabular}{lccccccccc} 
\Xhline{1px}
\multirow{2}{*}{\textbf{Model}}   &  \multicolumn{3}{c}{\makebox[0pt][c]{\hspace{1.6em}\textbf{\datasetnospace-Short}\hspace{-10.1em}\rule[-0.7ex]{11.8em}{0.5px}}}   & \multicolumn{3}{c}{\textbf{\makebox[0pt][c]{\hspace{1.6em}\textbf{\datasetnospace-Mid}\hspace{-9.5em}\rule[-0.7ex]{11.8em}{0.5px}}}}  & \multicolumn{3}{c}{\makebox[0pt][c]{\hspace{1.6em}\textbf{\datasetnospace-Long}\hspace{-9.8em}\rule[-0.7ex]{11.8em}{0.5px}}} \\
                                     
&  Visual      & CLIP         & TreeBLEU        &Visual       & CLIP         &  TreeBLEU       &  Visual     & CLIP         &   TreeBLEU       \\ 

\Xhline{0.7px}
LLaVA-v1.5-7B                        & $.43_{\pm.27}$  & $.60_{\pm.33}$  & $.07_{\pm.05}$  &$.21_{\pm.28}$ & $.29_{\pm.38}$   & $.05_{\pm.04}$  & $.19_{\pm.27}$ & $.28_{\pm.37}$ & $.04_{\pm.03}$   \\
LLaVA-onevision-0.5B                 & $.24_{\pm.31}$	& $.62_{\pm.11}$	& $.06_{\pm.03}$ & $.28_{\pm.31}$	& $.61_{\pm.10}$	& $.05_{\pm.03}$ & $.22_{\pm.29}$	& $.59_{\pm.11}$	& $.03_{\pm.02}$  \\
LLaVA-onevision-7B                   & $.34_{\pm.32}$	& $.63_{\pm.10}$	& $.08_{\pm.07}$ & $.30_{\pm.30}$	& $.64_{\pm.09}$	& $.06_{\pm.06}$ & $.30_{\pm.30}$	& $.61_{\pm.10}$	& $.04_{\pm.04}$  \\
CogAgent-Chat-18B                    & $.46_{\pm.31}$ & $.68_{\pm.11}$  & $.01_{\pm.03}$  & $.40_{\pm.31}$ & $.66_{\pm.10}$  & $.01_{\pm.02}$ & $.39_{\pm.30}$ & $.65_{\pm.10}$   & $.01_{\pm.01}$ \\
Gemini                               & $.35_{\pm.41}$	& $.75_{\pm.10}$	& $\bm{.16_{\pm.10}}$ & $.38_{\pm.40}$	& $.74_{\pm.11}$	& $\bm{.15_{\pm.08}}$ & $.34_{\pm.41}$	& $.73_{\pm.10}$	& $\bm{.14_{\pm.06}}$  \\
Claude                               & $.52_{\pm.43}$	& $.77_{\pm.10}$	& $.13_{\pm.08}$ & $.35_{\pm.42}$	& $.76_{\pm.09}$	& $.14_{\pm.08}$ & $.37_{\pm.43}$	& $.74_{\pm.09}$	& $.13_{\pm.06}$  \\
GPT-4V                               & $.68_{\pm.32}$  & $.74_{\pm.10}$ & $.12_{\pm.07}$  & $.65_{\pm.33}$  & $.71_{\pm.10}$ & $.11_{\pm.06}$  & $.62_{\pm.35}$ & $.67_{\pm.10}$  & $.10_{\pm.05}$  \\
GPT-4o                               & $\bm{.85_{\pm.16}}$  & $\bm{.77_{\pm.10}}$ & $.15_{\pm.09}$  & $\bm{.81_{\pm.20}}$ & $\bm{.77_{\pm.09}}$  & $.13_{\pm.08}$ & $\bm{.82_{\pm.18}}$ & $\bm{.74_{\pm.09}}$  & $.11_{\pm.05}$  \\
\Xhline{1px}
\end{tabular}
\label{tb_results_2}
\end{table*}

\subsection{Effectiveness of Training Dataset (RQ1)}
Table~\ref{tb_results_1} presents the performance of \modelname both on the WebSight and \dataset datasets, compared to other benchmark models on the WebSight dataset.
From this figure, we can observe that our method consistently outperforms all specialized baselines across all three metrics on the real-world test dataset, noting that these specialized models are fine-tuned on the WebSight dataset. Comparative experiments also demonstrate that the base model, Pix2Struct, achieves a significant performance boost when fine-tuned on our training dataset compared to WebSight. For TreeBLEU—a metric measuring the recall of 1-height subtrees in the target DOM tree—our approach surpasses both specialized and general-purpose models, indicating that our model better reflects real-world node types and substructures. Additionally, on the two visual similarity metrics—visual score and CLIP similarity—our model exceeds most general-purpose models and either matches or outperforms GPT-4V. Collectively, these results demonstrate that our dataset offers greater practical potential than synthetically generated datasets and suggest that our proposed training dataset can effectively unleash the potential of MLLMs in webpage generation.

\subsection{Benchmarking on the Test Datasets (RQ2)}
Table~\ref{tb_results_2} presents the benchmarking performance of several general-purpose MLLMs using the \dataset test dataset.
From this figure, we can observe several interesting findings:
\textbf{(1) Generating lengthy code is challenging.} Almost all metrics for nearly all models drop significantly as the target code length increases. For example, as the dataset transitions from \datasetnospace-Short to \datasetnospace-Mid and finally to \datasetnospace-Long, the highest TreeBLEU score for specialized models drops from 0.35 to 0.15, the highest CLIP similarity decreases from 0.73 to 0.69, and the highest Visual Score declines from 0.78 to 0.65.
\textbf{(2) Model size matters.} In LLaVA family, several models show a significant improvement across all metrics as model parameters increase, with LLaVA-v1.5-7B and LLaVA-onevision-7B achieving the best performance, while LLaVA-onevision-0.5B performs poorly across all metrics, indicating that MLLMs require more parameters to achieve better results in webpage generation tasks.
\textbf{(3) Most general-purpose MLLMs struggle with webpage code generation.} Among these models, only GPT-4V matches the performance of our model trained on \datasetnospace, while GPT-4o significantly outperforms all other models. All remaining general-purpose models generally underperform compared to specialized models, with consistently low scores across all metrics.

Notably, GPT-4o significantly outperforms all specialized and other general-purpose MLLMs across all metrics. Moreover, its performance remains highly stable as the complexity increases, without showing significant degradation. For instance, as the dataset transitions from \datasetnospace-Short to \datasetnospace-Mid and then to \datasetnospace-Long, the visual score changes from 0.85 to 0.81, and then to 0.82.
However, our goal is not to have small models outperform super MLLMs with hundreds of billions of parameters, but to assist MLLMs in webpage generation and enable smaller models to achieve competitive performance.

\section{Related Work}
\mypara{Image Representation Learning.}
Image representation learning is essential for various vision tasks, such as image-to-text tasks (e.g., image captioning~\cite{DBLP:conf/cvpr/VinyalsTBE15} and image-to-code), image classification~\cite{DBLP:journals/ijcv/RussakovskyDSKS15}, object detection~\cite{DBLP:conf/cvpr/GirshickDDM14}, and semantic segmentation~\cite{DBLP:journals/pami/ShelhamerLD17}. A number of studies have explored different methods and models for deriving meaningful latent vectors from images. For example, \citet{chen2020simple} investigated the use of contrastive learning to train image encoders on large-scale datasets. The ViT introduced a novel approach by dividing an image into fixed-size patches and using a Transformer to process them, which allows for handling variable resolutions~\cite{alexey2020image}. In addition, \textit{Diffusion Models} (DM)~\cite{DBLP:conf/nips/HoJA20,DBLP:conf/cvpr/RombachBLEO22} have demonstrated remarkable success in image representation learning, understanding, and generation tasks.

\mypara{Code Generation.}
Neural language models have made significant strides in code intelligence~\cite{wan2024deep}, encompassing tasks such as code summarization~\cite{wan2018improving,wang2020reinforcement}, code search~\cite{wan2019multi}, and code generation~\cite{bi2024iterative,DBLP:conf/kbse/Sun000J0L24}. Recently, with the advancement of LLMs in text generation, several specialized LLMs for code have been introduced, including CodeT5+~\cite{wang2023codet5+}, InCoder~\cite{fried2022incoder}, StarCoder~\cite{Li2023StarCoderMT}, Code Llama~\cite{roziere2023code}, WizardCoder~\cite{luo2023wizardcoder}, Qwen-Coder~\cite{hui2024qwen2}, and DeepSeek-Coder~\cite{guo2024deepseek}. 
These models have significantly transformed the software development landscape through various production tools, such as Copilot~\cite{GitHub2022copilot}, CodeWhisperer~\cite{Aws2022codewhisperer}, and Replit~\cite{Replit2022replit_ai}. In contrast to general-purpose LLMs designed for code generation, this paper focuses on the specialized task of code generation from webpage designs, a distinct domain within code generation.

\mypara{Image to Code.}
Several early works have made pioneering contributions to generating code from images. For instance, to reverse engineer source code from GUIs, \citet{Tony2018_pix2code} proposed pix2code, which was trained on a synthetic dataset to generate DSL code. Sketch2code~\cite{Alex2019_Sketch2code} explored two approaches for generating website code from wireframe sketches: a computer vision-based method that detects UI elements and structures, and a deep learning-based method. \citet{Wu2021ScreenPT} tackled the challenge of screen parsing by predicting UI hierarchy graphs from screenshots using Faster-RCNN~\cite{Ren2015FasterRT} to encode images and an LSTM-based attention mechanism to construct graph nodes and edges. Pix2Struct~\cite{Kenton2023_Pix2Struct}, pre-trained to generate simplified HTML from masked website screenshots, demonstrated significant improvements in visual understanding across nine tasks.
More recently, several powerful commercial models have emerged, including OpenAI's GPT-4V and GPT-4o~\cite{DBLP:journals/corr/abs-2303-08774}, Google DeepMind's Gemini~\cite{DBLP:journals/corr/abs-2312-11805}, and Anthropic's Claude~\cite{TheC3}. These models exhibit impressive performance across a wide range of tasks, including image understanding and code generation, with the advantage of continuous adaptation and optimization through interactive chat interfaces.
In contrast to previous studies, our research primarily focuses on proposing a comprehensive real-world dataset for both model training and testing purposes.

\section{Discussion}
\subsection{Practical Challenges to Study}
In the course of our research, we identify three practical challenges that need to be addressed to achieve the ideal generation of webpage code from design images. These challenges are presented here to guide future research:
\textbf{(1) Lengthy code generation.}
As shown in Table~\ref{tb_stat}, despite our efforts to clean up noise in the webpage code, such as invisible elements, the HTML text remains lengthy, reflecting its complexity to some extent. This presents significant challenges to both the training effectiveness and efficiency of MLLMs.
\textbf{(2) Capturing structural information of webpage designs.}
Given the potential overlap of sub-elements in images and the lack of distinct borders for some elements, extracting structured or hierarchical information from images presents a significant challenge. Our empirical study of GPT-4V reveals that, while it excels in capturing text and color from images when generating webpage code, it struggles with capturing the hierarchy of UI elements. 
\textbf{(3) Generation of image elements.}
All existing webpage code generation models fail to accurately reproduce the image elements in the designs, severely hindering their practical application. There is an urgent need for a framework capable of generating or extracting image elements from the original design and assembling them into the final webpage code.

\subsection{Limitations and Future Directions}\label{sec_limi}
Firstly, although we employ a carefully designed neural scorer to enhance data quality, this method inherently involves some degree of subjectivity and achieves approximately 90\% accuracy. As a result, some low-quality data may remain in the final dataset. However, we consider this acceptable given the trade-off between efficiency and quality, as manually screening millions of data points is impractical.
Secondly, the results in Table~\ref{tb_results_1} and Table~\ref{tb_results_2} show that all models exhibit significant variance across certain metrics. This indicates that the model's generation capability is not sufficiently stable and performs poorly on some test cases, highlighting the need for a more robust framework. Additionally, it underscores the importance of designing better evaluation metrics for code generation from webpage designs. In future work, we plan to assess the quality of generated code and webpages using LLMs~\cite{chen2024mllm}.
Finally, since our dataset is derived from web-crawled data, it may contain a minimal amount of inappropriate content, such as violent material, despite extensive filtering efforts.

\section{Conclusion} 
In this paper, we introduce \datasetnospace, the first large-scale, real-world dataset with layout information designed for generating webpage code from designs. The dataset contains over 2.56 million samples for both training and testing. We detail the dataset construction pipeline and provide an in-depth analysis of the curated data, which highlights its diversity. We then fine-tune an MLLM, \modelname, on our training dataset. Along with two visual metrics and a structural measure, we evaluate \modelname against other baselines on the proposed dataset. The experimental results show that our dataset significantly enhances the ability of MLLMs to generate code from webpage designs. We believe that this dataset and benchmark will contribute to advancing research in this field.

\begin{acks}
This work is supported by the Major Program (JD) of Hubei Province (Grant No. 2023BAA024).
\end{acks}

\balance
\bibliographystyle{ACM-Reference-Format}
\bibliography{ref}

\clearpage
\appendix

\begin{figure*}[h]
    \centering
    \footnotesize
    \begin{tcolorbox}[colback=white, colframe=gray, boxrule=0.5mm, title = Prompt for using GPT-4V in webpage generation task]
You are an expert Tailwind developer. You take screenshots of a reference webpage from the user, and then build single-page apps using Tailwind, HTML and JS.

- Make sure the app looks exactly like the screenshot.

- Make sure the app has the same page layout like the screenshot, i.e., the generated html elements should be at the same place with the corresponding part in the screenshot and the generated html containers should have the same hierachy structure as the screenshot.

- Pay close attention to background color, text color, font size, font family, padding, margin, border, etc. Match the colors and sizes exactly.

- Use the exact text from the screenshot.

- Do not add comments in the code such as "<!-- Add other navigation links as needed -->" and "<!-- ... other news items ... -->" in place of writing the full code. WRITE THE FULL CODE.

- Repeat elements as needed to match the screenshot. For example, if there are 15 items, the code should have 15 items. DO NOT LEAVE comments like "<!-- Repeat for each news item -->" or bad things will happen.

- For images, use placeholder images from https://placehold.co and include a detailed description of the image in the alt text so that an image generation AI can generate the image later. In terms of libraries,

- Use this script to include Tailwind: <script src="https://cdn.tailwindcss.com"></script>

- You can use Google Fonts

- Font Awesome for icons: <link rel="stylesheet"  href="https://cdnjs.cloudflare.com/ajax/libs/font-awesome/5.15.3/css/all.min.css"></link> Return only the full code in <html></html> tags.

- Do not include markdown "'''" or "'''html" at the start or end.
\end{tcolorbox}
    \caption{Prompt for using GPT-4V in webpage generation task.}
    \label{fig_example_for_pormpt}
\end{figure*}

\section{The Implementation of TreeBLEU}
\label{sec_treeblue}
The implementation of TreeBLEU includes two parts: the generation of a simplified HTML DOM tree and collecting 1-height subtrees.
\begin{itemize}[nolistsep, leftmargin=*]     
    \item Generation of a simplified HTML DOM tree.
    The original HTML DOM tree contains many element attributes that are not necessary for calculating the similarity of substructures. Therefore, we first use bs4~\cite{homepage_bs4} package in Python to parse the HTML text and obtain a DOM tree starting from the \texttt{<HTML>} node, which only includes the tag names of elements.
\begin{algorithm}[h]
\caption{Get All 1-height Subtrees of a DOM tree.}
\begin{algorithmic}[1]
    \Require A multiway tree node $root(childs,name)$
    \Ensure A set $S$ of all 1-height subtrees
    \State Initialize an empty set $S$
    \Function{Traverse}{$node,S$}
        \If{number of children of $node \neq 0$}
        \State Initialize an empty string $subtree$
        \State Append $node.name$ to $subtree$
        \For{each child $c$ in $node.childs$}
        \State Append $c.name$ to $subtree$
        \EndFor
        \State Add $subtree$ to $S$
        \EndIf
        \For{each child $c$ in $node.childs$}
        \State \Call {Traverse}{$c,S$}
        \EndFor
    \EndFunction
\State \Call {Traverse}{$root,S$}
\end{algorithmic}
\label{alg_subtrees}
\end{algorithm}
    \item Collecting 1-height subtrees. As shown in Algorithm~\ref{alg_subtrees}, We use the string sequence of a node's tag name and those of its child nodes to represent a 1-height subtree. By traversing the entire DOM tree and removing duplicates, we obtain the set of all 1-height subtrees.
\end{itemize}

\begin{figure}[H]
    \centering
    \includegraphics[width=0.95\linewidth]{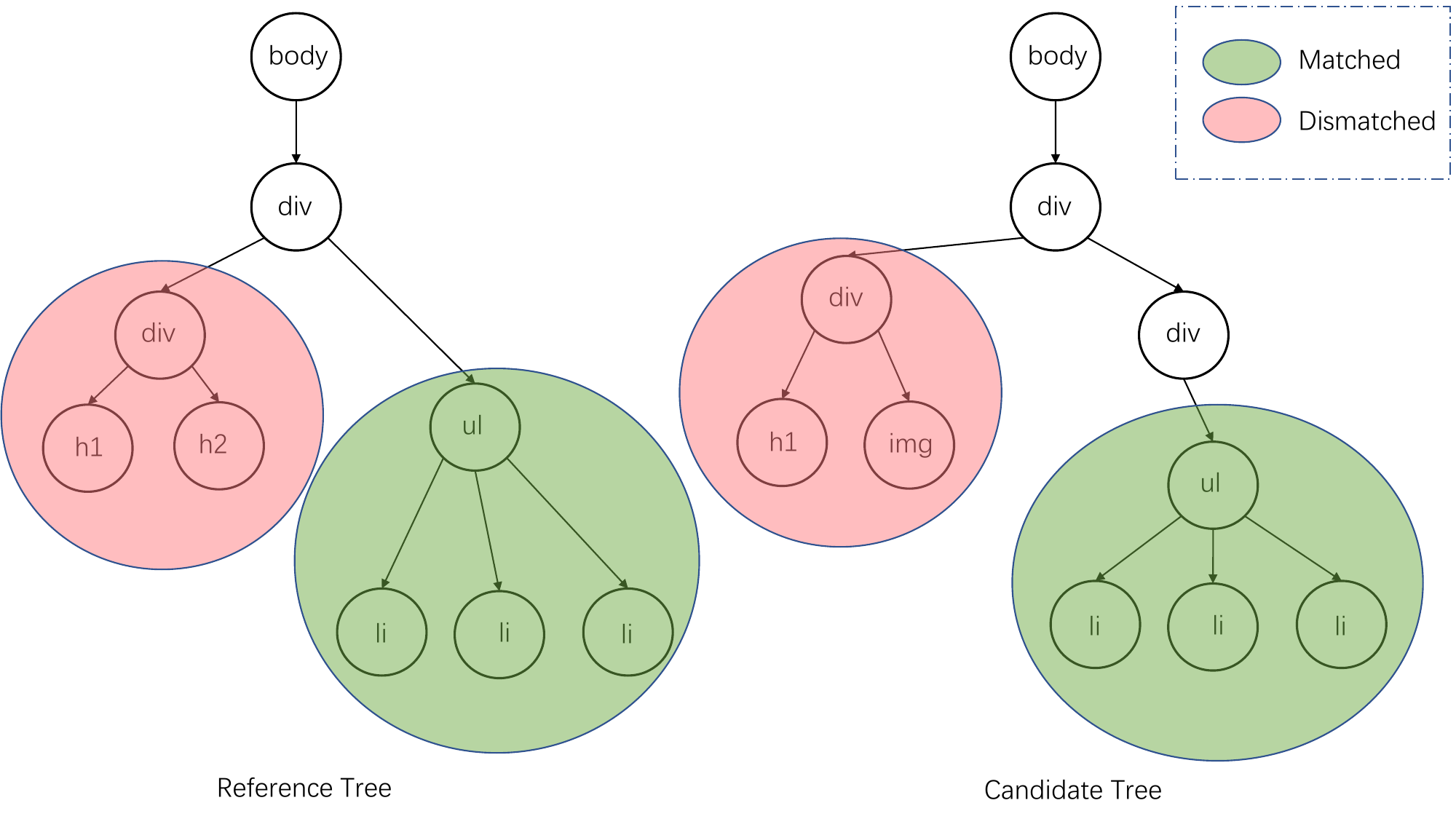}
    \caption{An example of subtree matching in TreeBLEU.}
    \label{fig_tree_matching}
    \vspace{-1em}
\end{figure}

Given that the subtree collections of the generated HTML DOM tree and the reference are represented as sets of strings, subtree matching can be straightforwardly performed through string comparison. The TreeBLEU value can then be calculated using Eq.~\ref{eq_treebleu}. 
An example of subtree matching in TreeBLEU is presented in Figure~\ref{fig_tree_matching}.

\section{More Data Statistics}
\label{sec_other_stat}
\begin{figure}[H]
     \centering
     \includegraphics[width=0.68\linewidth]{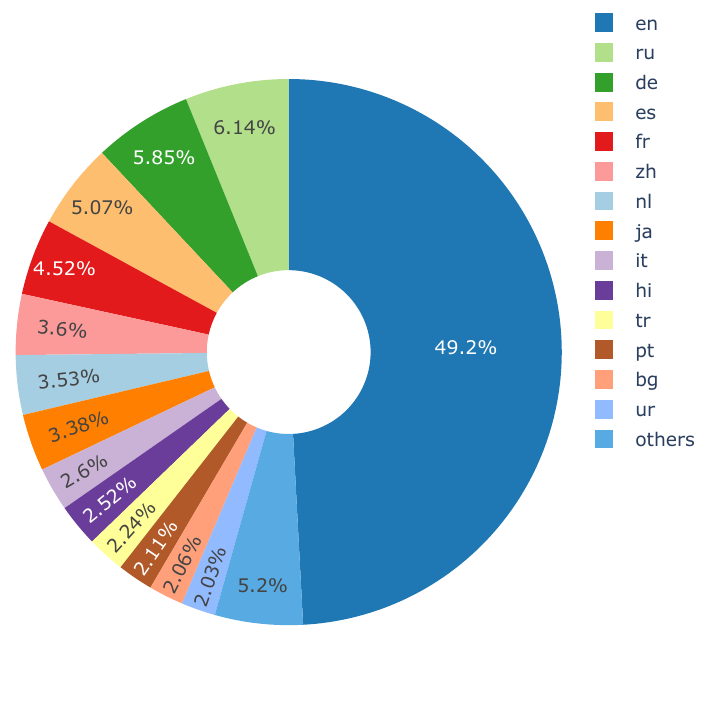} 
     \caption{Language distribution in the training dataset.}
     \label{fig_lang}
\end{figure}

We perform a comprehensive statistical analysis on the textual content of webpages, and the findings are illustrated in Figure~\ref{fig_lang}.
English constitutes approximately 50\% of the corpus, with other languages such as Russian, German, Spanish, and French each representing nearly 5\% of the total corpus. 

\section{Filtering Inappropriate Content}
\label{sec_harmfull_filtering}
The main steps and details of the cleansing process are as follows.

\mypara{Step 1: Filtering harmful images in screenshots.} We employ a widely used NSFW detector~\cite{homepage_nfsw} from Hugging Face, which classifies images as either ``norma'' or ``NSFW'' (\textit{Not Safe for Work}) with high accuracy. This detector is based on a  ViT model fine-tuned on a dataset containing both ``safe'' and ``explicit'' images. It predicts two scores: ``normal'' and ``NSFW'', with lower NSFW scores indicating a lower probability of harmful content. A conservative threshold of 0.04 is adopted, meaning only samples with NSFW scores below this value are retained.
    
\mypara{Step 2: Multi-language harmful keyword filtering.} We apply harmful keyword filtering to the web text using two popular GitHub repositories: bad words 1~\cite{homepage_badwords1} and bad words 2~\cite{homepage_badwords2}. The first list contains dirty, naughty, obscene, and otherwise inappropriate words in multiple languages (\textit{e.g.}, ``fuck''), while the second serves as a supplementary list with additional sensitive or stop words. Samples with more than 20 occurrences of these bad words are removed. While normal web pages may occasionally contain a small number of inappropriate words, those with excessive amounts of inappropriate content, such as adult websites, tend to have significantly higher word counts.

The thresholds for both the NSFW score and the frequency of bad words are determined through experiments on the first data chunk (see Figure~\ref{tb_nfsw_values}). We select thresholds that balance maximizing sample retention with minimizing the presence of harmful content. To evaluate the effectiveness of the filters, we manually review and document the filtering results for the first chunk under different thresholds. The table below clearly indicates that an NSFW threshold of 0.04 produces the best results.

\begin{table*}[!ht]
\centering

\arrayrulecolor{black}
\caption{The filtering performance across different NSFW threshold values.}
\begin{tabular}{ccccccccccccc} 
\Xhline{1px}
\multirow{2}{*}{Threshold} & \multirow{2}{*}{\begin{tabular}[c]{@{}c@{}}Total\\Num\end{tabular}} & \multicolumn{3}{c}{Removed by NSFW filter}                                                                                                        & \multicolumn{4}{c}{Retained after NSFW filter}                                                                                                                                                                          & \multicolumn{4}{c}{Retained after NSFW filter  bad word filter}                                                                                                                                                                                               \\ 
\cline{3-13}
                  &                                                                     & \multicolumn{1}{l}{\begin{tabular}[c]{@{}l@{}}total\\num\end{tabular}} & \multicolumn{1}{l}{\begin{tabular}[c]{@{}l@{}}good\\num\end{tabular}} & \begin{tabular}[c]{@{}c@{}}miskill\\ratio\end{tabular} & \begin{tabular}[c]{@{}c@{}}total\\num\end{tabular} & \begin{tabular}[c]{@{}c@{}}bad\\num\end{tabular} & \begin{tabular}[c]{@{}c@{}}retention\\ratio\end{tabular} & \begin{tabular}[c]{@{}c@{}}toxic\\ratio\end{tabular} & \begin{tabular}[c]{@{}c@{}}total\\num\end{tabular} & \begin{tabular}[c]{@{}c@{}}bad\\num\end{tabular} & \begin{tabular}[c]{@{}c@{}}retention\\ratio\end{tabular} & \begin{tabular}[c]{@{}c@{}}toxic\\ratio\end{tabular}                                       \\ 
\hline
0.02              & \multirow{5}{*}{1536}                                               & 891                                                                    & 883                                                                   & 57.49\%                               & 645                                                & 0                                                & 41.99\%                                 & \textcolor[rgb]{0,0,0}{0.00\%}                   & 645                                                & 0                                                & 41.99\%                                 & \textcolor[rgb]{0,0,0}{0.00\%}                                                         \\
0.03              &                                                                     & 324                                                                    & 318                                                                   & 20.70\%                              & 1212                                               & 2                                                & 78.91\%                                 & \textcolor[rgb]{0,0,0}{0.17\%}                   & 1210                                               & 0                                                & 78.78\%                                 & \textcolor[rgb]{0,0,0}{0.00\%}                                                         \\ 
\rowcolor{gray!10!white} 0              &                                                                     & 133                                                                    & \multicolumn{1}{c!{\color[rgb]{0,0,0}}}{128}            & \textcolor[rgb]{0,0,0}{8.33\%}                     & 1403                                               & 3                                                & \textcolor[rgb]{0,0,0}{91.34\%}                      & \textcolor[rgb]{0,0,0}{0.21\%}                   & 1400                                               & 0                                                & \textcolor[rgb]{0,0,0}{91.15\%}                      & \multicolumn{1}{c!{\color[rgb]{0,0,0}}}{\textcolor[rgb]{0,0,0}{0.00\%}}  \\ 
0.05              &                                                                     & 70                                                                     & 66                                                                    & \textcolor[rgb]{0,0,0}{4.30\%}                     & 1466                                               & 4                                                & \textcolor[rgb]{0,0,0}{95.44\%}                      & \textcolor[rgb]{0,0,0}{0.27\%}                   & 1463                                               & 1                                                & \textcolor[rgb]{0,0,0}{95.25\%}                      & 0.07\%                                                                    \\
1.00              &                                                                     & 0                                                                      & 0                                                                     & \textcolor[rgb]{0,0,0}{0.00\%}                     & 1536                                               & 8                                                & \textcolor[rgb]{0,0,0}{100.00\%}                     & 0.52\%                              & 1530                                               & 2                                                & \textcolor[rgb]{0,0,0}{100.00\%}                     & 0.13\%                                                                    \\
\arrayrulecolor{black}\Xhline{1px}
\end{tabular}
\label{tb_nfsw_values}
\end{table*}

\end{document}